\documentclass[runningheads, table, dvipsnames]{llncs}
\usepackage{graphicx}

\usepackage{tikz}
\usepackage{comment}
\usepackage{amsmath,amssymb} 
\usepackage{color}

\usepackage[accsupp]{axessibility}  

\usepackage{enumitem}

\usepackage[pagebackref=true,breaklinks=true,letterpaper=true,colorlinks,bookmarks=false]{hyperref}
\hypersetup{
    colorlinks=true,
    linkcolor=red,
    filecolor=magenta,      
    urlcolor=blue,
}
\usepackage{orcidlink}

\usepackage{ifthen}
\usepackage[normalem]{ulem}

\newcommand{\etal}{\emph{et~al.}}
\usepackage{xspace}
\newcommand{\MergeNet}{MergeNet\xspace}

\newcommand{\newdataset}{In-the-wild-RS\xspace}
\newcommand{\ourdb}[2]{\emph{\textbf{#1}} and \emph{\textbf{#2}}}

\usepackage{subcaption}
\begin{document}
\pagestyle{headings}
\mainmatter
\def\ECCVSubNumber{5951}  

\title{Combining Internal and External Constraints for Unrolling Shutter in Videos}

\titlerunning{Combining Internal and External Constraints for Unrolling Shutter in Videos}
\authorrunning{E. Naor \etal}
\author{
\small
Eyal Naor
\,
Itai Antebi
\,
Shai Bagon\orcidlink{0000-0002-6057-4263}
\,
Michal Irani
}
\institute{
\small 
Dept. of Computer Science and Applied Math,
The Weizmann Institute of Science\\
\textbf{\emph{Project Website:~}}\url{www.wisdom.weizmann.ac.il/\~vision/VideoRS}
}
\maketitle
\begin{abstract}
Videos obtained by rolling-shutter (RS) cameras result in spatially-distorted frames. These distortions become significant under fast camera/scene motions. Undoing effects of RS is sometimes addressed as a \emph{spatial} problem, where objects need to be rectified/displaced in order to generate their correct global shutter (GS) frame. However, the cause of the RS effect is inherently temporal, not spatial. In this paper we propose a space-time solution to the RS problem. We observe that despite the severe differences between their $xy$ frames,  a RS video and its corresponding GS video tend to share the exact same $xt$ slices -- up to a known sub-frame temporal shift. Moreover, they share the same
distribution of small 2D $xt$-patches, despite the strong temporal aliasing within each  video. This allows to constrain the GS output video using \emph{video-specific} constraints 
imposed by the RS input video. Our algorithm is composed of 3 main components:
(i)~Dense temporal upsampling between consecutive RS frames using an off-the-shelf method,
(which was trained on regular video sequences), 
from which we extract GS ``proposals’’. (ii)~Learning to correctly merge an ensemble of such GS ``proposals’’ using a dedicated \MergeNet.
(iii)~A \emph{video-specific} zero-shot optimization which imposes the similarity of $xt$-patches between the GS output video and the RS input video. Our method obtains state-of-the-art results on benchmark datasets, both numerically and visually, despite being trained on a  small 
synthetic RS/GS dataset. Moreover, it generalizes well to new complex RS videos with motion types outside the distribution of the training set (e.g., complex non-rigid motions) -- videos which competing methods trained on much more data cannot handle well. 
We attribute these generalization {capabilities to the combination of external and internal constraints.
\vspace*{-0.2cm}}
\end{abstract}


\begin{figure}[t]
    \centering
    
     \hspace{0.3cm} RS input   \hspace{1.0cm}     SUNet~\cite{fan_SUNet_ICCV21}  \hspace{0.9cm}  RSSR~\cite{fan_RSSR_ICCV21}  \hspace{1.2cm}   \textbf{\textcolor{red}{Ours}}   \hspace{0.6cm}   Ground Truth GS

    \includegraphics[width=\linewidth]{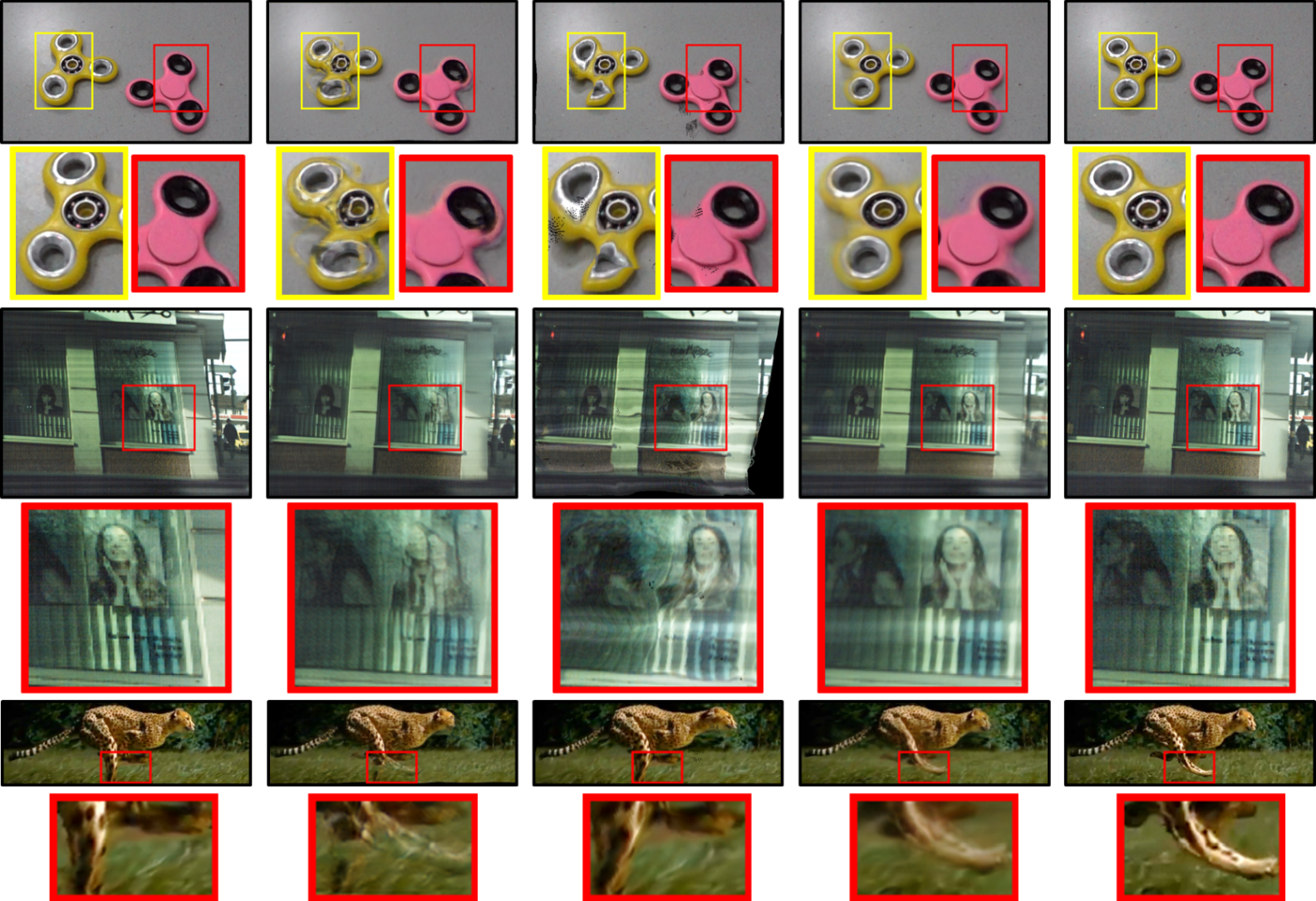}
    \vspace*{-6mm}
    \caption{\textbf{Examples of RS-induced distortions for various scene dynamics (and attempts to fix them).} {\it (Top) \underline{Rotational motion}: the round tip of the rotating pink spinner turns into  an  ellipse in the RS video,  and  its  position  is  displaced  within  the  frame. (Middle) \underline{Camera translation}: straight vertical lines get tilted in RS. (Bottom) \underline{Non-rigid motion}: the limbs of a fast running cheetah become completely distorted and dislocated (in a non-parametric way).
   SotA methods (SUNet~\cite{fan_SUNet_ICCV21}, RSSR~\cite{fan_RSSR_ICCV21}) 
    fail to generalize  to RS distortion types outside their \mbox{trainning set (especially non-rigid scenes), whereas our method does favorably.}}}
    \label{fig:teaser}
    \vspace*{-0.8cm}]
\end{figure}


\section{Introduction}

\begin{figure}[t]
    \centering
    \begin{subfigure}[b]{\textwidth}
        \caption{\raggedright \underline{Initial RS-GS Misalignment:}}
        \hspace{3.0cm} Frame \#11 \hspace*{1.5cm} Frame \#33 \hspace*{1.5cm} Frame \#77 \\
        \includegraphics[width=\linewidth]{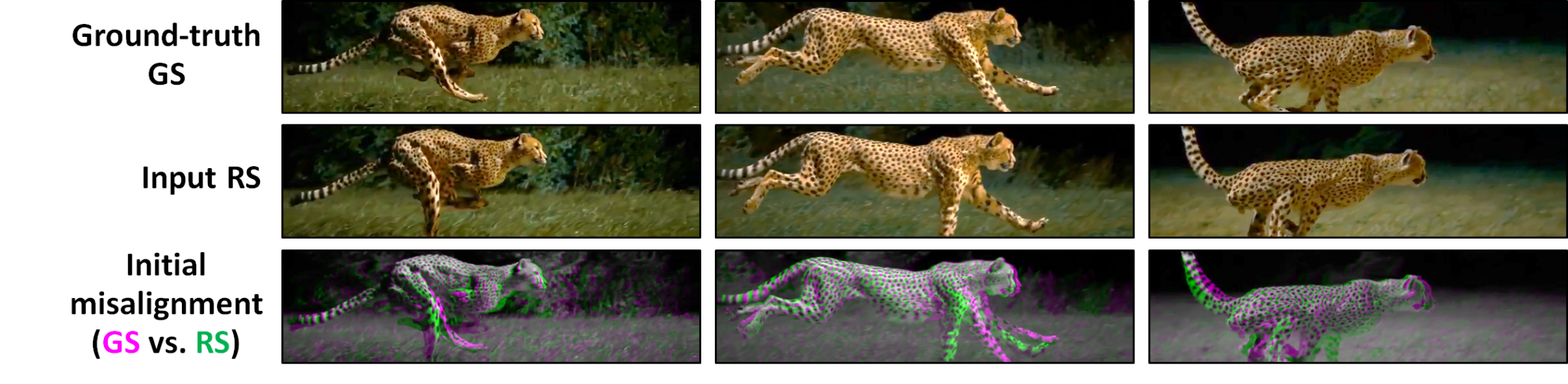}
        \label{fig:cheetah_top} 
    \end{subfigure}
    \vspace*{-3mm}
    \begin{subfigure}[b]{\textwidth}
        \caption{\raggedright \underline{Residual misalignemnts after fixing RS Effects:} \emph{(shown for Frame\#11)}}\vspace*{-1mm}
        \hspace{2.3cm} SUNet~\cite{fan_SUNet_ICCV21} \hspace{1.1cm} RSSR~\cite{fan_RSSR_ICCV21} \hspace{1.2cm} \textbf{\textcolor{red}{Ours} \hspace{0.9cm} Ground-Truth}
        \includegraphics[width=\linewidth]{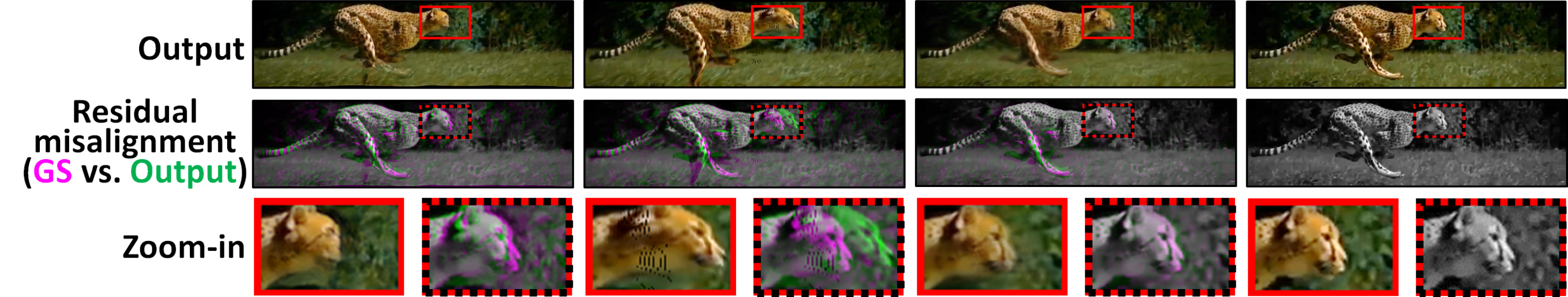}
        \label{fig:cheetah_bottom}
    \end{subfigure}
    \vspace*{-6mm}
    \caption{\textbf{Visualizing the RS-GS Misalignments.} {\it (a) Visualization of the initial distortion/misalignment between the RS frame (inserted into the G band) and its corresponding GS frame (inserted into the R\&B bands). Notice the complexity of the artifacts, especially in non-rigid areas. Grayscale indicates good alignment, whereas Green and Magenta indicate misalignment. {(b) Visualizing residual misalignment between reconstructed GS and ground-truth GS.} While state-of-the-art competitors fail to properly correct the distortions and position of the running cheetah (e.g., see zoom-in on face), our method does so favorably.}}
    \label{fig:cheetah}
    \vspace*{-0.6cm}
\end{figure}

Rolling shutter (RS) 
cameras are widely used in many consumer products.
In contrast to global shutter (GS) cameras, which capture all pixels of a single frame simultaneously, RS cameras capture the image pixels row by row. 
Consequently, a variety of spatial distortions (e.g., tilt, stretch, curve, wobble) appear under fast camera/scene motion.
Examples of such distortions can be seen in Fig.~\ref{fig:teaser} (e.g., the round hole at the tip of the rotating pink spinner turns from a circle into an ellipse, and its position is displaced within the frame).
Fig.~\ref{fig:cheetah}  exhibits the degree of misalignment between temporally corresponding RS and GS frames in a video of a fast running cheetah.

RS correction methods can be broadly classified as either \emph{single-frame} \cite{lao2018robust,rengarajan2017unrolling,rengarajan2016bows,zhuang2019learning} or \emph{multi-frame}~\cite{albl2020two,fan_RSSR_ICCV21,fan_SUNet_ICCV21,liu2020deep,ringaby2012efficient,vasu2018occlusion,zhuang2017rolling,zhuang2020image}. Attempting to reconstruct a GS frame from a single RS frame is highly ill-posed, as it does not exploit the inherent temporal aspect of the RS problem. Single-frame methods thus require significant assumptions on either the camera motion (pure translation, pure rotation, etc.) or  on the scene  (planar scene, straight lines, etc.) 
As a result, the current leading methods~\cite{fan_RSSR_ICCV21,fan_SUNet_ICCV21} are multi-frame ones.
These are  the methods we compare against.

In this paper we propose a space-time solution to the RS problem.
The RS problem is fundamentally temporal, since it stems from different rows being captured at different times. 
In fact, the RS frame captures the ``correct" (undistorted) image rows, but at the wrong times. 
Thus, despite the severe spatial distortions between RS and GS frames, we observe that a RS video and its corresponding GS video share the exact same $xt$ slices -- up to a known sub-frame temporal shift. This observation is invariant to the types or complexity of the camera/scene motions.
We thus repose the problem of rectifying RS videos as a problem of correctly shifting and interpolating 
$xt$ slices. 

\begin{figure}[t]
    \centering
    \includegraphics[width=\textwidth]{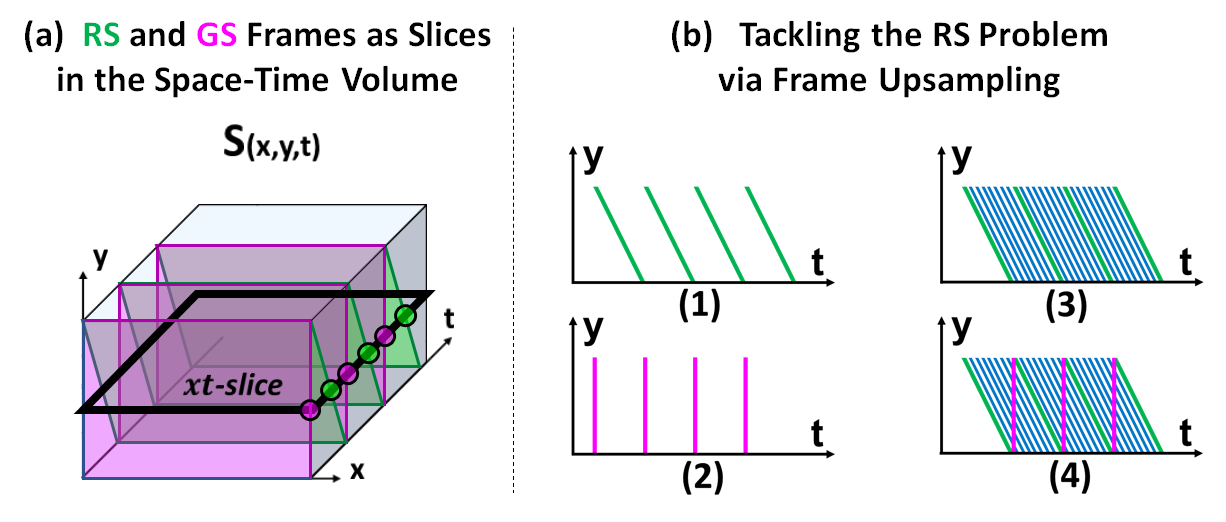}
    \vspace*{-0.7cm}
    \caption{\textbf{Space-time relations between RS and GS.} \textbf{(a)} {\it RS and GS frames in the Space-Time Volume $S(x,y,t)$. In magenta:  GS frames record all pixels simultaneously, thus capturing $xy$ planes in  $S(x,y,t)$.  In green:  RS frames record row-by-row, capturing ``slanted" planes in $S(x,y,t)$. The GS and RS videos share the same $xt$ slices, up to a known sub-frame shift.}  \textbf{(b)} {\it Re-casting the RS problem as a temporal frame-upsampling problem. 
    (1)~RS input frames are displayed as green slanted lines in a side  $yt$-view  of  $S(x,y,t)$. 
    (2)~GS frames are displayed as vertical magenta lines in the side view.
    (3)~Temporal frame-upsampling  ``fills" the space time volume by generating intermediate RS frames. (4)~Sampling the relevant row from each interpolated RS frame allows to reconstruct the GS frames.}}
    \label{fig:3Dvol_and_TI}
    \vspace*{-0.25cm}
\end{figure}

More specifically, our algorithm is composed of 3 steps:
(i)~We use an off-the-shelf frame interpolation algorithm~\cite{DAIN} (pre-trained on regular videos)
to densely fill the space-time volume between consecutive RS frames.
Were the temporal-upsampling perfect, recovering the GS frame 
would then be trivial 
-- simply sample the correct row from each interpolated video frame (see Fig.~\ref{fig:3Dvol_and_TI}(b)).
(ii)~Since general-purpose frame interpolation methods are prone to errors (and more so in presence of RS effects),  
we apply the temporal-interpolation algorithm to multiple augmentations of the RS video, to generate multiple GS ``proposals".
We train a small RS-specific \MergeNet to correctly merge an ensemble of such GS “proposals”, while being sensitive to local RS idiosyncrasies. 
Due to its simplicity, it suffices to train \MergeNet on a  \emph{small and synthetic} dataset of RS/GS video pairs.
This is in sharp contrast to competing SotA methods~\cite{fan_RSSR_ICCV21,fan_SUNet_ICCV21}, which train a different model for each new dataset.
(iii)~Finally, we 
observe that a RS/GS video pair tends to share the same distribution of small 2D $xt$-patches.
We use this observation to 
impose \emph{test-time video-specific} constrains 
on the $xt$-patches of the GS output video (to match those of the  RS input video). 

Our method thus benefits from both  Internal and External constraints: on the one hand, we utilize a frame interpolation method~\cite{DAIN}, \emph{externally} trained on large datasets of videos, 
while on the other hand, our zero-shot test-time optimization takes advantage of \emph{internal} video-specific distribution of $xt$-patches.

Our method outperforms existing RS video methods on a large variety of RS video types -- evaluated both on existing RS benchmark datasets, as well as on a new dataset we collected of challenging videos with highly non-rigid motions (which are lacking in existing benchmark datasets). Particularly, our method outperforms prior methods \emph{\textbf{by a large margin}}  on RS videos of complex non-rigid scenes. We attribute the  generalization capabilities of our algorithm to its combination of external and internal constraints.

\vspace*{0.1cm}
\noindent Our contributions are thus several fold: \\
$\bullet$ We re-cast the RS correction problem as a temporal upsampling problem. As such, we can leverage advanced frame interpolation methods (which have been pre-trained on a large variety of real-world videos).\\
$\bullet$  We observe that a RS video and its corresponding GS video share the same small $xt$-patches, despite significant temporal aliasing exhibited in both videos. This allows {to impose \emph{video-specific} constraints on the GS output, \emph{at test-time}}. \\
$\bullet$ We curated and released a new dataset of RS/GS video pairs, which pushes the envelope of RS benchmarks to include also complex non-rigid  motions. \\
$\bullet$ We provide state-of-the-art results on RS video benchmarks. 


\section{Related Work}

RSSR~\cite{fan_RSSR_ICCV21} and SUNet~\cite{fan_SUNet_ICCV21} are the SotA methods, and are also the most closely related to our approach. SUNet~\cite{fan_SUNet_ICCV21}  introduced symmetric consistency constraints  while warping consecutive RS frames to produce a single in-between GS frame. They use a constant velocity motion model, and train a network to convert the optical flow between the RS frames to produce ``RS undistortion-flow''. RSSR~\cite{fan_RSSR_ICCV21}  inverts the mechanism of RS in order to recover a high framerate GS video from consecutive RS frames. 
Both methods have  complex networks that require much training data. In fact, they have a different trained model for each dataset. This limits their performance and generalization capabilities on new out-of-distribution RS videos (see Figs.~\ref{fig:teaser},\ref{fig:cheetah}). 
Furthermore, since \cite{fan_RSSR_ICCV21} rectifies the frames using RS undistortion-flow, the results contain holes on occlusion boundaries.
In contrast, our method 
uses a single light network (trained on a synthetic dataset) on all benchmarks, with leading performance,
while exhibiting good generalization capabilities on new out-of-distribution RS videos.

A number of other previous Deep-Learning multi-frame methods  
were also proposed. 
In~\cite{Liu_2020_CVPR} an array of networks was trained to warp and predict a GS frame from two consecutive RS frames. They further provide 2 benchmark datasets of paired RS/GS videos for training and evaluation. These datasets were later used by~\cite{fan_RSSR_ICCV21,fan_SUNet_ICCV21}, showing SotA results. We too experiment  on these datasets, and compare to the leading methods~\cite{fan_RSSR_ICCV21,fan_SUNet_ICCV21}. 
Two other nice recent RS methods,  but less closely related to us (as they require different input types) have been recently presented.  The method of~\cite{zhong2021towards} solves for a GS output from a \emph{blurry} RS input, whereas~\cite{albl2020two} proposes a method which uses two simultaneous RS cameras mounted on a single platform to produce one GS output.

\section{Inherent Relations between GS \& RS Videos}
\label{sec:GS-RS-relations}

\begin{figure}[t]
    \centering
    \includegraphics[width=\linewidth]{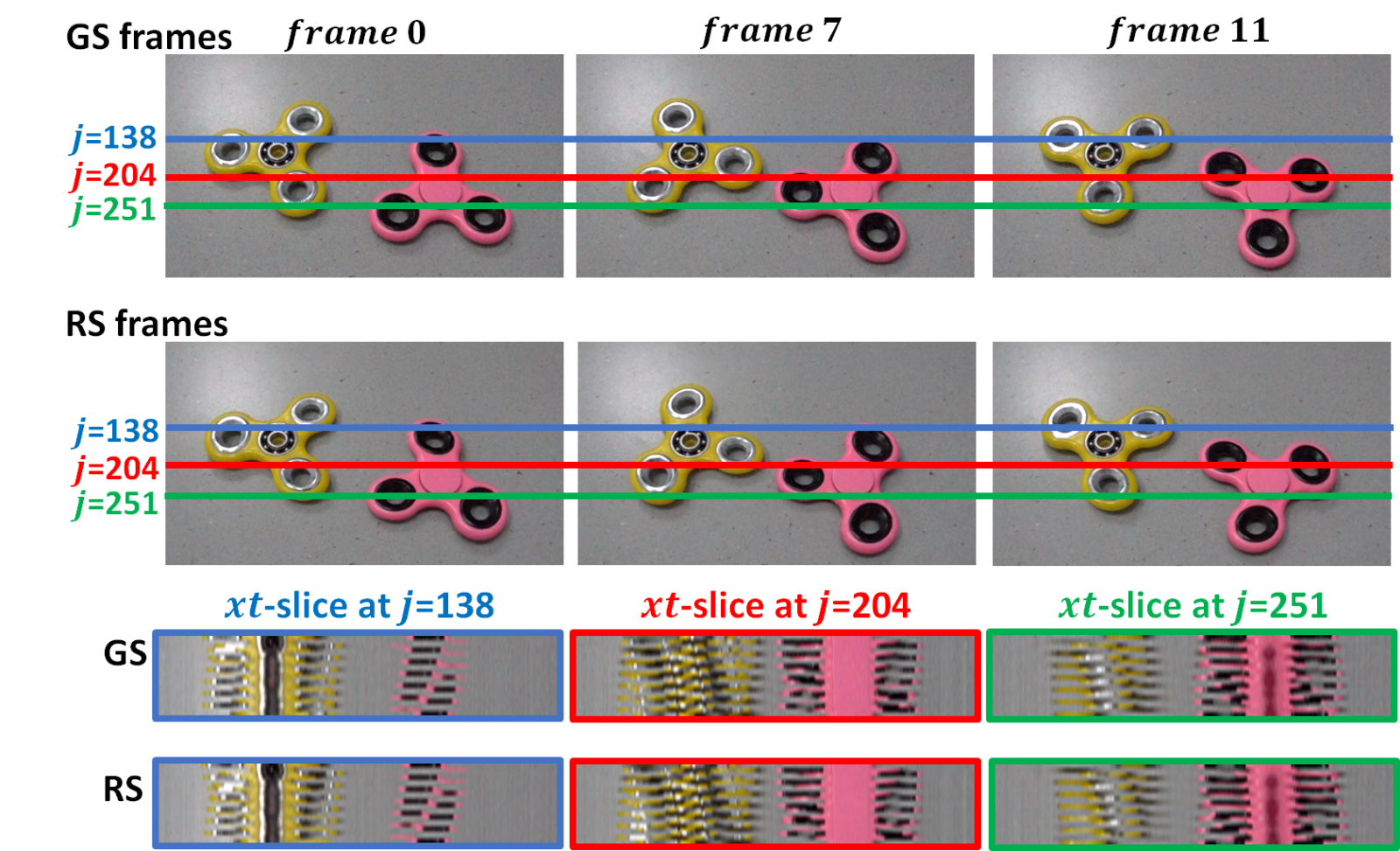}
    \caption{
    \textbf{GS-RS video pairs share the same $xt$-slices.} 
    {\it Corresponding \mbox{$xt$-slices}  
    \mbox{from the RS and GS video, at a few rows $j$ (= y). Although}
   { the RS frames exhibit strong distortions compared to their corresponding GS frames (circles turn into ellipses; angles of the spinner arms appear different),} corresponding  $xt$-slices (at same row~$j$) are very similar, as they sample the same $xt$-plane in the continuous space-time volume $S(x,y,t)$ at a 1D `sub-pixel' shift {in the $t$-direction (the vertical axis in those slices).} See Sec.\ref{sec:GS-RS-relations} and  Fig.~\ref{fig:3Dvol_and_TI}a for more details.}}
    \label{fig:xt_recurrence}
    \vspace*{-0.5cm}
\end{figure}


\noindent{\bf Claim:}
The $xt$-slice of the RS video is a \emph{shifted version}  of the $xt$-slice of the GS video, at the same row $j$, at a ``sub-pixel'' shift of $j/N$$<$$1$ along the $t$-axis, where $N$ is the number of rows in each frame. \\
%
\noindent{\bf Proof:}
Let $S(x,y,t)$ denote the continuous space-time volume (Fig.~\ref{fig:3Dvol_and_TI}a).
It is defined in the camera coordinate system, hence any camera motion relative to the scene can be regarded as scene motion relative to a static camera.
{Let $GS(i,j,k)$ and $RS(i,j,k)$ denote the GS and RS videos, respectively}, where $(i,j)$ are integer pixel coordinates, and $k$ is the  frame number, taken at time gaps of $\Delta T$ along the $t$ axis (w.l.o.g. we define $\Delta T=1$).
\emph{{By definition}},  RS and GS videos are { just different space-time samplings of} the continuous $S(x,y,t)$ as follows: 
\vspace*{-0.3cm}
\begin{equation}\label{eq:rs_gs}
GS(i,j,k) = S(i,j,k)  \  \ \ \text{and} \ \ \  RS(i,j,k)=S\left(i,j,k\!+\!\frac{j}{N}\right)
\end{equation}
\vspace*{-0.5cm}

\noindent
{Note that the above `sampling' equation explicitly  entails that, for any row $j_0$ ($j_0$=1,..N), \ $GS(i,j_0,k)$ and $RS(i,j_0,k)$ are related by a fixed 1D shift of $j_0/N$$<$$1$ along the $t$-axis, {$\forall i,k$}. These are exactly the $xt$-slices of the GS and RS videos, at row $j_0$. 
This proves our claim. $\blacksquare$ \\
%
\noindent
Although unintuitive, note that this observation, which is simply derived from Eq.(\ref{eq:rs_gs}), is {independent} of the content of the space-time volume $S$, hence is \emph{invariant to the type of scene/camera motions} (whether rotation, translation, etc).}\\
\noindent
{{More intuitively}:} The \emph{spatial frames} of the GS and RS videos are \emph{different planes} within the  space-time volume  $S(x,y,t)$ (see purple and  green slices in Fig.~\ref{fig:3Dvol_and_TI}a). Hence they naturally have very different appearances under severe camera or scene motions. 
{However, these RS distortions are manifested only in the $y$ direction (since every row $y$$=$$j$ is sampled at different time), but are not expressed in the $xt$ slices of $S$.}
The $xt$-slices  of GS and RS videos at row $j$
are just different samples of the \emph{same {shared} plane} (the black plane  in Fig.~\ref{fig:3Dvol_and_TI}a). Their samples are marked by purple and green points inside the black $xt$-plane. 
{Fig.~\ref{fig:xt_recurrence} exemplifies this phenomenon, displaying 3 corresponding $xt$~slices of a GS-RS video pair of a complex dynamic scene (fast rotating spinners).} Our algorithm for undoing RS effects {builds on top of this simple yet powerful observation, in 2 ways:}
\vspace*{-0.2cm}
\begin{enumerate}[leftmargin=*, wide=0pt]
\item
In principle, given this observation, the solution to the RS problem seems trivial: just back-warp each $xt$ slice of the RS video by its \emph{known} sub-frame temporal shift $j/N$ (determined by its row index $j$). However, such sub-frame warping is far from being trivial in the presence of \emph{temporal aliasing}, which is very characteristic of video data (due to low frame-rate compared to fast scene/camera motions).
To address this issue, we resort to a pre-trained state-of-the-art temporal upsampling/interpolation network~\cite{DAIN} (which was trained on a huge collection of regular videos). 
However, video temporal upsampling methods have their own inaccuracies, and even more so on RS videos (which are video types they 
{were not} trained on). In Sec.~\ref{sec:MergeNet} we propose 
an approach 
to address this issue (with very few RS-GS training data).
\item
We further employ in our algorithm another inherent relation of a RS-GS video pair: 
These 2 videos \emph{share the same pool of small $xt$-patches}. It was shown in~\cite{shahar2011space} that small 3D space-time patches (e.g., 7$\times$7$\times$3) recur abundantly within a \emph{single} natural video of a dynamic scene. This space-time recurrence is an inherent property of the continuous dynamic world.  
Moreover, such patches were shown to appear in different aliased forms at different locations within the video (which gave rise to temporal super-resolution from a single video~\cite{shahar2011space,zuckerman2020across}).
The RS distortions affect the $y$ direction of small space-time patches; however, these distortions do not affect the $xt$ direction of these patches. 
Thus, while the GS and RS videos may not share the same small 3D space-time patches, they do share the same small 2D $xt$-patches (e.g., 7$\times$3), despite the temporal aliasing. 
The continuous version of a 2D $xt$-patch will appear many times in  the continuous dynamic scene, hence will appear at multiple $xt$ slices in each video, each time sampled at a different sub-frame (``sub-pixel'') temporal shift. Therefore, despite the temporal aliasing, each small $xt$-patch in the GS video will likely 
have similar patches (with the appropriate sub-frame temporal sampling) within some $xt$~slices of the RS video. \\
We empirically measured the strength of  GS-RS \emph{cross}-video recurrence of {$xt$-patches}, compared to their \emph{internal}-recurrence within the GS video itself. This was estimated as follows:
We randomly sampled a variety of GS-RS video pairs, which cover a variety of different motion types (rotation, translation, zoom, and non-rigid motions). 
For each 7$\times$3 $xt$-patch $p(x,t)$ in each GS video, we computed 2 distances: (i)~the distance to its nearest-neighbor (NN) $xt$-patch in the GS video \  \mbox{$d_{GS}(p)$$=$$\lVert p-\text{NN}(p,GS) \rVert$}, and (ii)~the distance to its nearest-neighbor $xt$-patch in the paired RS video \ \mbox{$d_{RS}(p)$$=$$\lVert p-\text{NN}(p,RS) \rVert$.}  
We then measure for each patch the ratio $r$$=$$\frac{d_{RS}(p)}{d_{GS}(p)}$, which tells us how worse is the patch similarity \emph{across} the 2 videos compared its similarity \emph{within} the GS video. Our empirical evaluations show that $mean(r)$$=$$1.13$, i.e., on average, the cross GS-RS patch distance $d_{RS}(p)$ is only $\times 1.13$ larger than the internal patch distance $d_{GS}(p)$. This holds not only for smooth patches, but also for patches with high gradient content (which correspond to sharp edges and high temporal changes). In fact, when measured only for the \emph{top 25\% of $xt$-patches with the highest gradient magnitude}, for 61\% of them  $r \leq 1.1$, and for 85\% of them  $r \leq 1.5$. This indicates high similarity of small $xt$-patches between the 2 videos.\\
We employ this observation to impose an additional \emph{video-specific} prior on our GS output video, at test-time, constraining the output by the the collection of 7$\times$3 $xt$-patches of the RS input video (see Sec.~\ref{sec:ItaiNet} for more details).
\end{enumerate}

\section{Method}\label{sec:method}

Our algorithm is composed of 3 main steps. 
First, we use an off-the-shelf general-purpose frame-upsampling algorithm, in order to extract the relevant rows from different temporally-interpolated RS frames, and compose them into GS frames (Sec.~\ref{sec:temp-interp}).
To compensate for the fact that the frame-upsampling algorithm is general-purpose (not RS-specific), we apply this process repeatedly on several different augmentations of the RS input video,  which result in several ``GS proposals" per frame.
In the second step, we train and use a RS-specific  \MergeNet, to merge the GS proposals into a coherent GS frame.  Since this is a simple network with few layers and a very narrow receptive field, it suffices to  train it with a small synthetic RS-GS dataset (Sec.~\ref{sec:MergeNet}).
Lastly, we use a zero-shot approach to refine the resulting GS video to adhere to the patch statistics of the input RS video, \mbox{thus reducing blurriness and other undesirable visual artifacts (Sec.~\ref{sec:ItaiNet}).}

{We note that these 3 main steps are separate as they are trained at different times, \emph{on different types of data:} \ 
\emph{Stage1} - leverages off-the-shelf SotA
{frame-interpolation}
methods,  pretrained on large datasets of \emph{general} videos. \ 
\mbox{\emph{Stage2}~- MergeNet} is trained on RS videos, {at train-time}.} \ 
\emph{Stage3} - is applied to the specific test video, {at test-time}.

\begin{figure}[t]
    \centering
    \includegraphics[width=.95\linewidth]{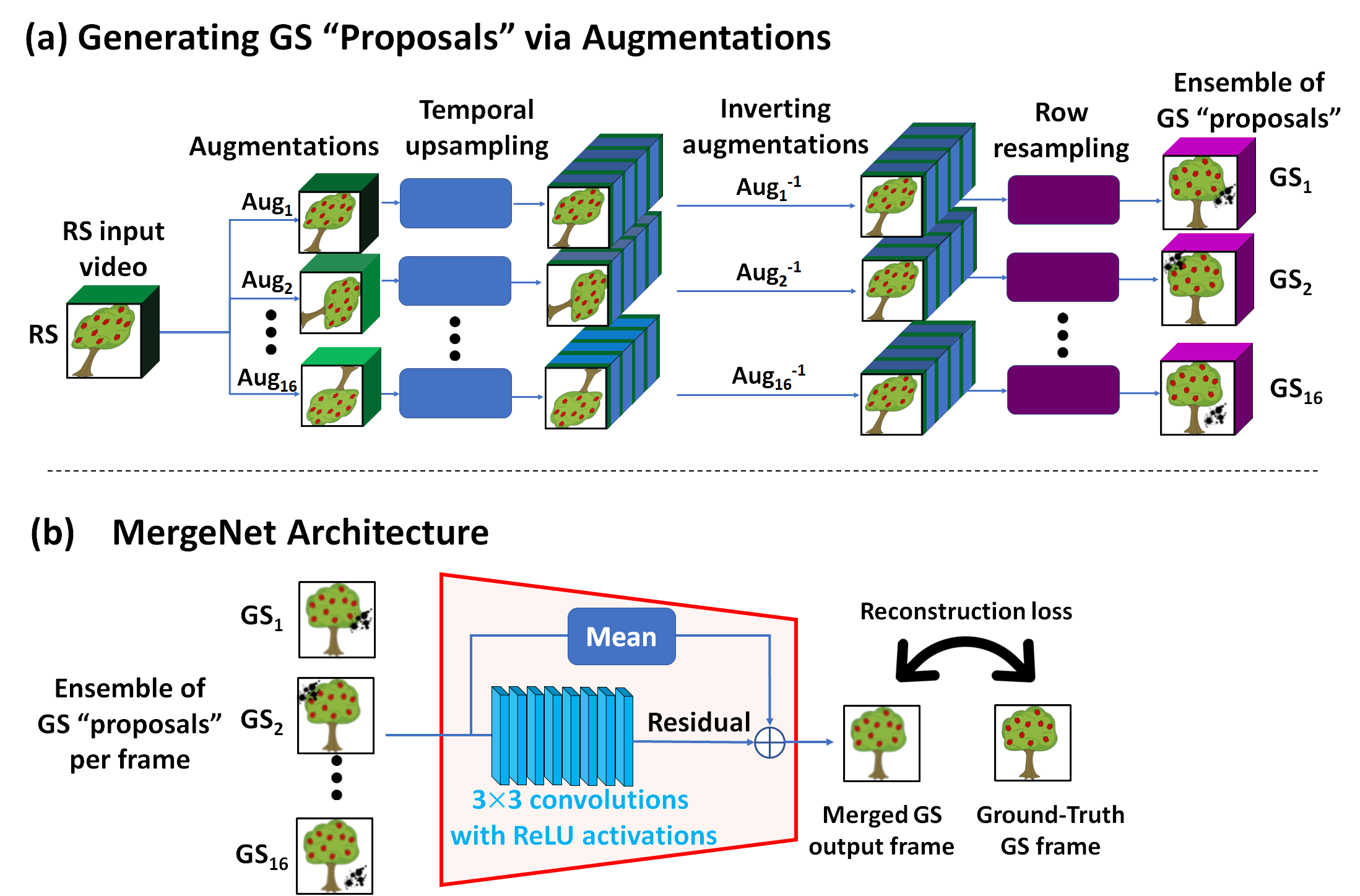}
    \vspace*{-0.25cm}
    \caption{\textbf{\MergeNet.} {\it   
    (a)~We adapt a \underline{general-purpose} frame-upsampling method~\cite{DAIN} (trained on regular videos) to temporally-upsample RS videos, by applying it to multiple augmentations of the RS video. This generates 16 GS ``proposals'' per frame (see Sec.~\ref{sec:MergeNet}). 
    (b)~We train a small RS-specific ``MergeNet'' to correctly merge an ensemble of  GS frame ``proposals'' into a single coherent GS frame. 
  \MergeNet learns the residual correction w.r.t. the mean of the 16 frame proposals, 
  while being sensitive to local RS idiosyncrasies.
   Being a rather shallow CNN (8~layers of 3$\times$3 convolutions followed by ReLU activations), with a small (17$\times$17) receptive field,
      it suffices to train \MergeNet on a small and synthetic dataset of RS/GS video pairs. 
      All hidden layers are with 64 channels.}}
\label{fig:mergenet}
    \vspace*{-0.5cm}
\end{figure}

\subsection{Generating GS Proposals via Temporal Frame-Upsampling}
\label{sec:temp-interp}
As we observe in Sec.~\ref{sec:GS-RS-relations}, GS $xt$-slices can be recovered by 
shifting the RS \mbox{$xt$-slices} at each row $j$ by a ``subpixel'' (subframe) shift of size $j/N$. 
However, such sub-frame warping is challenging due to severe temporal-aliasing, which is very characteristic of video data (regardless of whether a RS or a GS camera was used).
To address this issue, we resort to a state-of-the-art  off-the-shelf  temporal frame interpolation network, DAIN~\cite{DAIN}, which was
trained on a large in-the-wild video dataset~\cite{xue2019vimeo90k} depicting a wide range of motions and complex scene dynamics.
DAIN works by estimating a robust depth-aware flow between consecutive frames and utilizes this flow to efficiently perform temporal frame upsampling to an arbitrary (user-defined) framerate.
Using the same flow to interpolate all in-between frames makes DAIN's output temporally consistent, reducing flickers and other undesired artifacts in its predictions (regardless of the temporal interpolation rate).

Since each row in the RS frame comes  from a different temporal offset, the number of interpolated frames between every two RS input frames is determined by the number of rows in each frame. 
That is, for a RS video with $N$ rows/frame we need to temporally-upsample $\times$$N$ the original frame-rate, producing   $N$$-$$1$ additional frames between every 2 input RS frames (typical values of $N$ are on the order of hundreds of rows).
Once interpolated, we compose a \emph{GS proposal} frame by taking the relevant row from each temporally-interpolated RS frame, as illustrated in Fig.~\ref{fig:3Dvol_and_TI}(b). 
However, since DAIN did not train on videos with noticeable RS distortions, the interpolated frames of DAIN {on RS data} are often imperfect, affecting the quality of the GS proposal.
This problem is addressed using the second step of our algorithm, described next.

\subsection{\MergeNet: Merging Multiple GS Proposals}
\label{sec:MergeNet}

DAIN~\cite{DAIN} is general-purpose frame-interpolation method, trained on many videos (not RS-specific). To make better use of DAIN  on RS videos, which may contain distortions and dynamic behavior outside its training distribution, we apply DAIN on several different augmentations of the RS input video,  resulting in several different ``GS proposals" per frame.
We apply the following augmentations to the input RS video (spatial and temporal augmentations, which ignore the RS scanning order):  reversing the video in time (play backwards), spatially rotating it by $k\cdot90^\circ$ ($k$$=$$1,2,3,4$), and spatial horizontal flipping. This results in 16 pre-determined augmentations in total.
DAIN is then applied to temporally upsample each of these 16 augmented videos, followed by inverse augmentation  and appropriate row-subsampling, to generate 16 ``GS proposals'' (Fig.~\ref{fig:mergenet}(a)).

The resulting GS proposals are not identical: for some videos DAIN performs better on several of the augmentations but not on others, depending on motions and distortions specific to each frame.
Therefore, it is crucial to merge these proposals in a non-trivial manner to ignore regions with unwanted artifacts in proposals and taking advantage of better recovered GS regions. 
To that end we use a RS-specific ``\MergeNet'' to combine these proposals into a coherent GS frame. 
Note that although motions and distortions in videos may be \emph{globally} complex, they are still characterized by \emph{locally} simple linear motions.
Therefore, we deliberately design \MergeNet with 
a small receptive field ($17\!\times\!17$ pixels), to learn how to merge and fix small patch-wise GS-proposals.
Consequently, a limited synthetic video dataset with a variety of simple global affine motions provides enough diversity of locally-linear ones. These offer sufficient examples to train \MergeNet to learn how to correctly merge and fix small patch-wise GS-proposals. 
We train \MergeNet on a small available synthetic  dataset of affine-induced RS/GS videos pairs (the ``Carla-RS'' train-set of~\cite{Liu_2020_CVPR} -- see Sec.~\ref{sec:datasets}).
Although  trained on local patch-wise synthetic  examples, 
\MergeNet generalizes well to real complex RS videos of highly non-rigid scenes.

To conclude, we reduce the difficult task of correcting RS-distorted frames, to a much simpler task of adapting a generally-applicable frame interpolation algorithm to handle RS videos. The ``heavy-lifting" global motions considerations are done by a general-purpose frame interpolation method, while the local adaptation to the idiosyncrasies of RS is done by our small \MergeNet.
Fig.~\ref{fig:mergenet}(b) shows the architecture of \MergeNet.

\subsection{Imposing \emph{Video-Specific} Patch Statistics at test-time}
\label{sec:ItaiNet}
The final step of our algorithm makes use of our observation that small $xt$-patches are shared by a RS and GS video of the same dynamic scene (see Sec.~\ref{sec:GS-RS-relations}). 
We thus constrain the $xt$-patches in our GS output video, to be from the same $xt$-patch distribution as the RS input video.
This is obtained via a short \emph{test-time} optimization over the $7\!\times\!3$ $xt$-patches of the  GS frames predicted from our previous \MergeNet step. 
This process changes the predicted GS patches to have smaller distances to their nearest-neighbor (NN) patches in the input RS video, while not allowing them to deviate too far from their initial predicted value.
We formulate this via the minimization of the loss function: $\mathcal{L}_{NN} + \lambda\cdot \mathcal{L}_{validity}$
\vspace*{-0.2cm} 
{\small\begin{eqnarray*}
\text{where} \qquad \qquad \mathcal{L}_{NN}& = & \sum_{ijk}\alpha_{ijk}\left\|GS_{patch_{ijk}} - RS_{patch_{NN[ijk]}}\right\|^2 \\ 
\mathcal{L}_{validity} & = &\sum_{ijk}(1-\alpha_{ijk})\left\|GS_{patch_{ijk}} - GS_{patch_{ijk}}^{initial}\right\|^2
\end{eqnarray*}}
%
\noindent
$\mathcal{L}_{NN}$ incorporates the distance of each $xt$-patch in location $i,j,k$ to its NN patch in the input RS video. Minimizing $\mathcal{L}_{NN}$ brings the predicted GS patches closer to those of the RS, thus improving the recurrence of $xt$ patches between the input RS and the output GS.
$\mathcal{L}_{validity}$ measures the distance between the current GS prediction and the initial output of \MergeNet, $GS^{initial}$. Minimizing $\mathcal{L}_{validity}$ helps stabilize the optimization process.
Following~\cite{MosseriZontak2013combining}, we give higher weight $\alpha_{ijk}$ to patches with strong edges, as their NNs are more reliable and less prone to over-fitting noise (see discussion on PatchSNR in~\cite{MosseriZontak2013combining}). Accordingly, the values of $\alpha_{ijk}\in[0, 1]$ are determined by Canny edge responses~\cite{canny1986computational} (computed on the $xt$-slices of the predicted GS output from \MergeNet), to weigh the patches accordingly.
We set $\lambda$ so that the 2 loss terms are of the same order of magnitude.

\section{Experimental Results}
\label{sec:experiments_and_results}
We validate our approach both quantitatively and qualitatively on existing RS/GS benchmark datasets, as well as on a new challenging dataset we curated.

\begin{figure}[t!]
    \centering
  \mbox{\hspace{2cm} RS input   \hspace{0.8cm}     SUNet~\cite{fan_SUNet_ICCV21}  \hspace{0.6cm}  RSSR~\cite{fan_RSSR_ICCV21}  \hspace{0.5cm}   \textbf{\textcolor{red}{Ours}}   \hspace{0.5cm}   GroundTruth GS}

    \hspace*{-0.5cm}\includegraphics[width=1.05\textwidth]{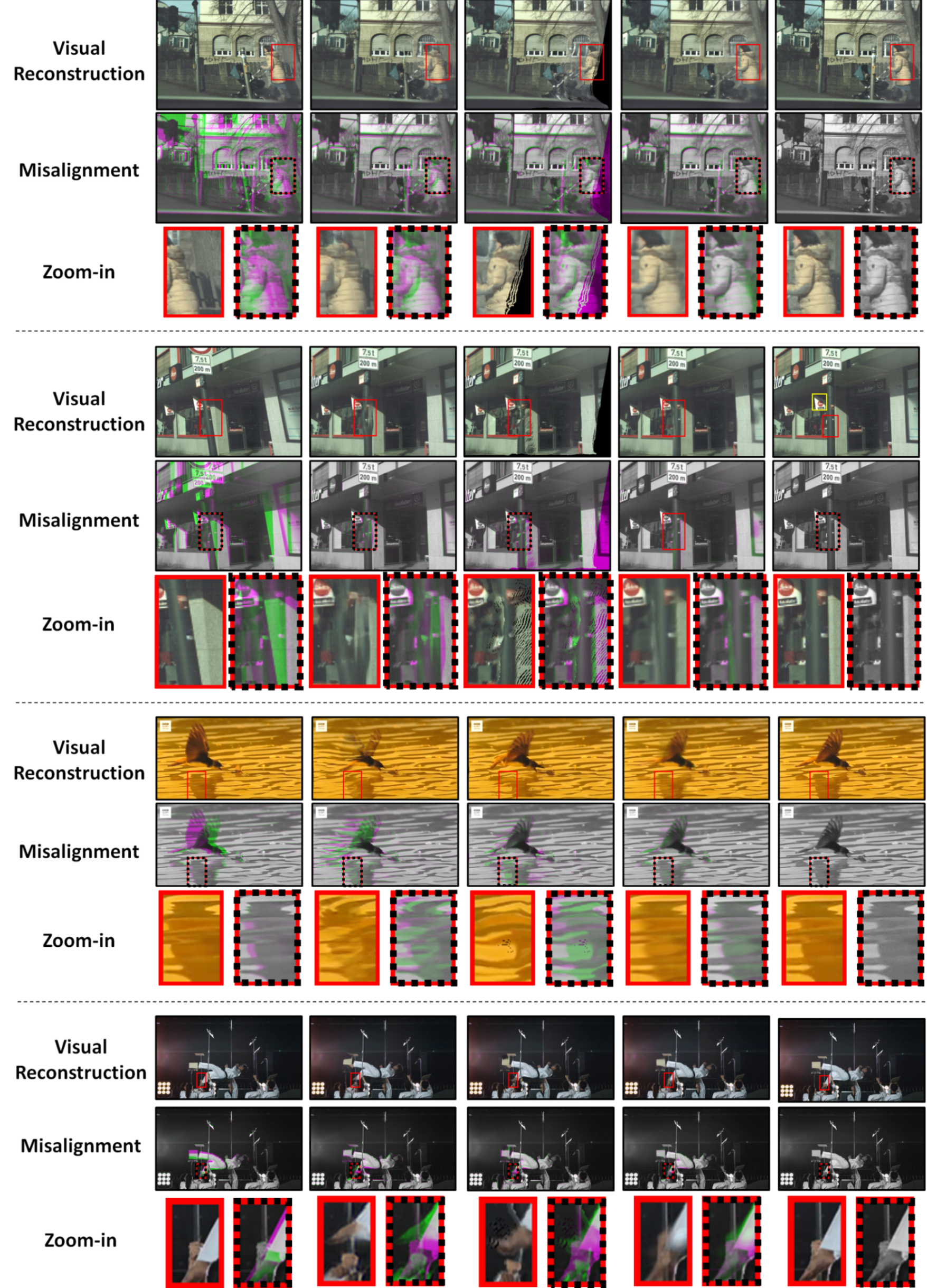}
    \caption{\textbf{Visually comparing reconstructions and residual misalignments.}} 
    \label{fig:image_page}
\end{figure}

\subsection{Datasets}
\label{sec:datasets}
RS  benchmark datasets with ground-truth GS data are comprised of aligned RS/GS video pairs.
Curating such datasets is technically challenging: 
one needs to capture or synthesize a very high framerate video, and then sub-sample it (vertically or diagonally -- see Fig.~\ref{fig:3Dvol_and_TI}) at $1/N$ of the framerate in order to generate aligned RS/GS videos (where $N$ is the number of rows per frame). 
Existing benchmark datasets  are thus relatively small ($\sim$2K frames), with a very limited variety of dynamic motions.

\noindent
$\bullet$ \textbf{Fastec-RS~\cite{Liu_2020_CVPR}:}
This dataset was created using a high-speed camera mounted on a driving car. Consequently, the motions are mostly horizontal translation, and the RS distortions are mostly affine ones. 
\emph{Fastec-RS} dataset comprises 76 sequences \mbox{with at most \emph{\textbf{34 frames}} per sequence~-- \emph{\textbf{56 Train-set, 20 Test-set}}.}

\noindent
$\bullet$ \textbf{Carla-RS~\cite{Liu_2020_CVPR}:}
This dataset was synthesized using the Carla simulator~\cite{dosovitskiy2017carla}; a virtual 3D environment. 
The virtual environment allows to simulate more complex camera motions; thus, Carla-RS, albeit synthetic, contains a wider variety of RS artifacts.
\emph{Carla-RS} comprises 250 sequences with \emph{\textbf{10 frames each}}~-- \emph{\textbf{210 Train-set, 40 Test-set}}. 
Being synthetic, this dataset further comes with occlusion maps between consecutive RS frames (regions which can potentially be ignored when evaluating the GS reconstruction results). Masks are used to calculate numerical results in \textit{Carla-RS masked}, which are also reported in Table~\ref{tab:numerical}.

\noindent
$\bullet$ \textbf{\newdataset [NEW]:} 
Neither \emph{Fastec-RS} nor \emph{Carla-RS} contain 
real complex non-rigid  motions, and as such are quite limited. 
To mitigate this lacuna, we curated \emph{\newdataset}, generated from 15 Youtube videos captured with high speed GS cameras, featuring complex non-rigid scene motions (running animals, flying birds, turbulent water, rotating spinners, etc), captured with unrestricted camera motions. 
For a few of these videos, we further generated  2-3 versions of GS/RS pairs, with varying degree  of RS complexity for the same scenes. 
On average, \emph{\textbf{~30 frames}} per sequence (some longer, some shorter)~-- \emph{\textbf{NO~Train-set, 15 Test-set}}. 
This dataset can be accessed through our \href{https://www.wisdom.weizmann.ac.il/~vision/VideoRS}{project page}.

\begin{table}[!t]
\vspace*{0.1cm}
    \centering
    \hspace*{-0.4cm}\begin{tabular}{|c||r||c|c|c|}
        \hline
        Dataset & &\quad \textcolor{red}{\textbf{Ours}} \quad\ & SUNet \cite{fan_SUNet_ICCV21} & RSSR \cite{fan_RSSR_ICCV21} \\
         \hline 
         \hline
        \textbf{Fastec-RS}~\cite{Liu_2020_CVPR}   & PSNR [dB] , SSIM \hspace*{0.01cm}& \textbf{28.573} , \textbf{0.8436} &  28.249 , 0.8277 & 21.175 , 0.7649\\
         \hline
         \hline
        \textbf{Carla-RS}~\cite{Liu_2020_CVPR}   & PSNR [dB] , SSIM \hspace*{0.01cm}& \textbf{31.430} , \textbf{0.9187} & 29.170 , 0.8499 & 24.776 , 0.8661\\
         \hline
         \hline
        \textbf{Carla-RS masked}& PSNR [dB] , SSIM \hspace*{0.01cm}& \textbf{31.840} , \textbf{0.9187} & 29.269 , 0.8499 & 30.137 , 0.8661\\
         \hline
         \hline
         \textbf{\newdataset$^{*}$}   & PSNR [dB] , SSIM \hspace*{0.01cm}& \textbf{27.919} , \textbf{0.900} &  24.531 , 0.8262 & 24.163 , 0.8376\\
         \hline   
    \end{tabular}
    \vspace*{0.1cm}
    \caption{\textbf{Numerical Evaluation:} {\it Our method outperforms~\cite{fan_SUNet_ICCV21,fan_RSSR_ICCV21} on all three benchmarks.
    Note that SUNet and RSSR trained \emph{benchmark-specific} models for Fastec-RS and Carla-RS, while our method uses a \emph{single} trained model for all datasets.
    $^{*}\,$\newdataset has no training set, hence for SUNet and RSSR we evaluated both their models, and reported \emph{their best result}. 
As can be seen, existing methods struggle to generalize to RS video types which are beyond those explicitly represented in their training sets. In contrast, our method (trained only on Carla-RS), generalizes much better (with a significant margin of \textbf{$+3.39~dB$}) on the challenging \newdataset.}}
    \label{tab:numerical}
    \vspace*{-0.8cm}
 \end{table}

\subsection{Quantitative Results}

Table~\ref{tab:numerical} shows PSNR and SSIM results of our reconstructed GS frames for the test sets of all three benchmarks.
We used only the synthetic training set of \mbox{\emph{Carla-RS}} 
\mbox{to train our
\MergeNet, and used the same network  for \emph{all} 3 benchmarks.}

We compare our results to the state-of-the-art methods SUNet~\cite{fan_SUNet_ICCV21} and RSSR~\cite{fan_RSSR_ICCV21}. 
These methods trained a \emph{different instance of their network} for each dataset -- \emph{Fastec-RS} and \emph{Carla-RS} (as opposed to our single \MergeNet).
Furthermore, since \emph{\newdataset} has no training set, we ran both  trained models of~\cite{fan_SUNet_ICCV21,fan_RSSR_ICCV21}, and reported \emph{\textbf{the  best performing one}} on \emph{\newdataset}'s test-set for each method (in both cases, it was the network trained on the synthetic Carla-RS dataset that performed best on \emph{\newdataset}).
Nevertheless, our method significantly outperforms SUNet and RSSR on all three benchmarks. 

It is interesting to see that although we did not use the training set of \mbox{Fastec-RS} at all, we outperform the models trained specifically on that benchmark, by \ourdb{+0.32 dB}{+7.4 dB}.
On Carla-RS, we outperform the competing models by \ourdb{+2.26 dB}{+6.7 dB}. 
More importantly, our method significantly outperforms SUNet and RSSR when evaluated on the challenging \newdataset dataset (with complex non-rigid motions, and no training set), by \ourdb{+3.39 dB}{+3.76 dB}. All in all, existing methods have more difficulty generalizing to new \mbox{types of RS distortions that are outside the distribution of their training set.}

Note that the numerical results of RSSR~\cite{fan_RSSR_ICCV21} are low due to the holes in their predicted GS frames, where no pixels were warped to by their undistortion-flow. 
The synthetic Carla-RS  further comes with  GT masks on occluded pixels, allowing RSSR to compare only on non-occluded pixels. 
These results are shown in the third row of Table~\ref{tab:numerical}. RSSR performs significantly better on \emph{non-occluded} pixels, surpassing SUNet. However, our method still performs better than both (\ourdb{+2.57  dB}{+1.7 dB}).

\vspace*{-0.3mm}
\paragraph{Ablation:} 
We further used Fastec-RS benchmark to evaluate the contribution of each step in our method. 
We note that most of the ``heavy lifting" comes from applying DAIN. This first step already yields good PSNR of 27.67~dB. 
Applying \MergeNet further improves results by additional $\sim$1~dB. 
Our last \emph{video-specific} test-time optimization step 
improved 30\% of the sequences by $\sim$0.2~dB, while yielding a smaller improvement on the other sequences.

\vspace*{-0.1cm}
\subsection{Qualitative Results}
\label{sec:results-qualitative}

Figs.~\ref{fig:teaser},\ref{fig:cheetah},\ref{fig:image_page} show visual results and comparisons.
Rectifying RS frames requires not only reconstructing good visual quality,  but no less important -- achieving good alignment w.r.t. the ground-truth GS frames.
This is a difficult  non-trivial task, as shown in Fig.~\ref{fig:cheetah}.
To better highlight the degree of \emph{residual misalignment} between the predicted GS frames and the ground-truth ones, we use the following visualization:
We convert the ground-truth and predicted GS frames to grayscale images. We place the  ground-truth RS in the red and blue channels, and the predicted  GS in the green channel.
Properly rectified areas are gray in the new visualization, while green or magenta highlighted areas indicate misalignments.
Figures~\ref{fig:cheetah} and~\ref{fig:image_page} use this visualization to highlight how our method better rectifies the scene in a variety of complex motion types. 
Compare, for example, our  proper alignment of the fast-moving sign-pole in the middle of Fig.~\ref{fig:image_page};
the non-rigid motion of the water ripples, the foot of the flipping man at the bottom of  Fig.~\ref{fig:image_page}, or the cheetah head in Fig.~\ref{fig:cheetah}.
Moreover, note our reconstruction of the round shape and position of the fast rotating spinner, as well as the hind leg of the cheetah, in Fig.~\ref{fig:teaser}.
These are a few examples of complex dynamic scenes from our new  \newdataset dataset.


\vspace*{-0.25cm}
\section{Limitations}
\vspace*{-0.25cm}
While current methods for temporal video-upsampling are quite advanced, 
this still forms the main bottleneck of our method. 
Our performance is bounded by the  limitations of  SotA frame-interpolation methods, 
which currently cannot handle videos with severe motion aliasing.
For example, a video recording of an extremely fast rotating propeller (much faster than the camera framerate), will appear in the video to be rotating in the reversed/wrong direction  (even when recorded by a GS camera). Current temporal interpolation methods cannot undo such severe motion aliasing, thus fail to generate the correct intermediate frames.
Our method fails when the frame-interpolation method breaks down. 
However, since the frame-upsampling is a standalone module in our method, it can be replaced as SotA frame-interpolation methods improve, leading to an immediate improvement in our algorithm, at no extra cost or effort.


\vspace*{-0.25cm}
\section{Conclusion}
\vspace*{-0.25cm}
We re-cast the RS  problem as a temporal frame-upsampling problem. As such, we can leverage advanced frame interpolation methods (which have been pre-trained on a large variety of complex real-world videos).
We bridge the gap between frame-interpolation of general videos to frame-interpolation of RS videos using a dedicated \MergeNet. We further observe that a RS video and its corresponding GS video share the same small $xt$-patches, despite significant temporal aliasing exhibited in both videos. This allows to impose video-specific constraints on the GS output, at test-time.  Our method obtains state-of-the-art results on a variety of benchmark datasets, both numerically and visually, despite being trained only on a small synthetic RS/GS dataset. Moreover, it generalizes well to new complex RS videos containing highly non-rigid motions  -- videos which competing methods trained on more data cannot handle well.

\vspace*{-0.25cm}
\paragraph*{\bf Acknowledgments:}
Project received funding from: the European Research Council (ERC grant No 788535), 
the Carolito Stiftung, and the D. Dan and Betty Kahn Foundation. 
Dr. Bagon is a Robin Chemers Neustein AI Fellow.

\clearpage


\bibliographystyle{splncs04}
\bibliography{thesis.bib}
\end{document}